# Adaptive mine planning under geological uncertainty: A POMDP framework for sequential decision-making

Hamza Khalifi [a], Jef Caers [b], Yassine Taha [a], Mostafa Benzaazoua [a], Abdellatif Elghali [a]

[a] Geology & Sustainable Mining Institute (GSMI), University Mohammed VI Polytechnic (UM6P), 43150, Benguerir, Morocco

[b] Department of Earth and Planetary Sciences, Stanford University, Stanford, CA 94305, USA

Correspondence to: Hamza Khalifi (hamza.khalifi@um6p.ma)

**Abstract**

Strategic mine production scheduling under geological uncertainty is conventionally formulated as a stochastic optimization problem in which a fixed extraction sequence and routing decisions are computed ex ante. This plan-driven paradigm treats uncertainty as passive: decisions are hedged across geological scenarios, but planning does not anticipate how future observations will inform future decisions. We propose a different perspective by formulating mine scheduling as a Partially Observable Markov Decision Process (POMDP), in which extraction and routing decisions are made sequentially with planning explicitly integrating the expectation of future belief updates.

To achieve computational tractability, we introduce a hybrid SA-POMDP architecture that combines simulated annealing-based (SA) value approximation with ensemble-based belief updating via ensemble smoother with multiple data assimilation (ES-MDA). At each decision epoch, candidate actions are evaluated through their expected long-term value under the current belief, and the belief is updated as mining observations are assimilated. This yields an adaptive policy rather than a fixed plan.

We evaluate the framework on a copper-gold open-pit mining complex with multiple processing destinations. Under a statistically consistent prior, the SA-POMDP reduces the expectation-reality gap from 22.3% to 4.6%, improving realized NPV by $8.4M relative to one-shot stochastic optimization. Under systematic prior misspecification of ±10%, the adaptive framework outperforms static planning by up to $44.6M (36.9%), demonstrating structural robustness beyond scenario hedging. These results show that sequential belief updating transforms geological uncertainty from a passive constraint into an active component of value creation.

## 1. Introduction

Strategic mine production scheduling is a primary economic driver of mineral resource management. It seeks to determine when, where, and how mineral resources should be extracted over the life of a mine to maximize economic value while satisfying operational, geological, and financial constraints. At its core, this involves the joint optimization of two coupled subproblems: block extraction sequencing and material routing (cut-off grade selection). Because these decisions dictate the temporal evolution of the project, they carry substantial long-term implications for the resilience and profitability of mining operations.

Traditionally, this problem is solved using a block model representation of the orebody, where thousands of discrete units are sequenced over decadal horizons. In this framework, the planner must balance immediate revenue against long-term accessibility, respecting spatial precedence- where overlying blocks must be removed to reveal deeper blocks- and annual capacity limits. Crucially, in modern high-complexity environments, cut-off grade behavior is no longer a fixed input; it emerges endogenously as the system reacts to the interaction between sequencing, processing constraints, and the inherent uncertainty of the resource [1,2].

Over several decades, the field has evolved from deterministic plans to sophisticated stochastic optimization frameworks. These current state-of-the-art methods- often utilizing mixed-integer programming or metaheuristics- have significantly improved projects realism by incorporating geological uncertainty, complex mining operations with multiple processing destinations, and waste management, etc. [3]. By optimizing across a range of simulated orebodies, these approaches aim to produce robust schedules that perform well on average.

However, a critical disconnect remains. While current methodologies acknowledge that geology is uncertain, they treat that uncertainty as passive. They assume that a robust plan can be determined ex ante and then managed through reactive adjustments rather than through policies explicitly conditioned on evolving geological beliefs. This ignores a fundamental physical reality: mining is a sequential sensing process. Every block extracted reveals precise grade information that should not only update the plan but redefine the underlying decision logic. By failing to formally condition future decisions on the expectation of learning, current frameworks treat time primarily as a discounting index rather than an information axis (value arising from belief evolution and policy adaptation), leaving the strategic value of information largely unexploited. As a result, observations obtained during mining- such as measured grades- do not systematically alter the set of geological hypotheses considered during optimization, nor do they change how future decisions are evaluated.

In current mine planning literature and industrial practice, uncertainty is typically represented through multiple geological realizations generated prior to optimization. Planning decisions are optimized against these realizations [4]. The resulting solution is a plan- a fixed block sequence and destination policy- that is subsequently executed and periodically revised.

Recent work has extended this paradigm to rolling-horizon frameworks that update geological simulations as mining progresses and re-optimize the production schedule at each decision epoch [5,6]. While such approaches represent an important step toward incorporating feedback from operations, they remain fundamentally plan-driven: at each epoch, a new schedule is generated by solving an optimization problem, and the first-period action is extracted from this solution. The decision logic is therefore implicit- embedded in the structure of the optimization- rather than represented as an explicit mapping from the current state of knowledge to an action.

A key limitation of periodic re-optimization is that it cannot fully undo irreversible commitments made in earlier periods. Decisions locked in when uncertainty was maximal remain fixed even as new observations reveal that alternative choices would have been preferable. Re-optimization at epoch $t$ is constrained by all prior extraction decisions, limiting the extent to which newly acquired information can improve outcomes. In this sense, while rolling-horizon approaches incorporate learning through updating and re-optimization, they do not fully exploit the sequential decision structure inherent in the problem.

A related but often overlooked limitation concerns how performance is interpreted. Conventional stochastic mine planning evaluates economic performance through distributions of net present value (NPV) across hypothetical geological realizations. While such distributions capture geological variability, they primarily reflect differences in input geology rather than the quality of the decision-making process itself. A high NPV in one realization occurs because that realization happens to be favorable, not because the decisions are inherently superior. Conversely, a low NPV reflects unfavorable geology rather than necessarily poor planning. By contrast, frameworks that explicitly model sequential learning evaluate policies whose expected value integrates over all possible observation sequences, measuring decision quality rather than scenario quality.

Mining is, by nature, a sequential and information-revealing process. Each extracted block provides grade measurements that reduce uncertainty about the surrounding orebody. Due to spatial correlation, these observations propagate beyond the mined block, altering expectations about future extraction opportunities. As mining progresses, some geological hypotheses become implausible while others gain credibility.

Yet in conventional formulations, the essential structure of sequencing decisions is committed when geological uncertainty is maximal, and no decisions are formally conditioned on how uncertainty is evolved through observation. This creates a fundamental asymmetry between the timing of commitment and the timing of information revelation: decisions are locked in precisely when they should be most provisional.

From a decision-theoretic perspective, this limitation arises because conventional formulations cannot express conditional strategies of the form "take action A now in order to determine whether action B should be taken later." In such formulations, the informational value of mining actions- their ability to reduce uncertainty and reshape future decisions- is implicitly assumed to be zero. This assumption is not the result of insufficient algorithms, but of posing strategic mine planning as a static optimization problem rather than as a sequential decision problem under partial observability.

It is important to note that this limitation is not resolved by increasing the number of geological realizations. Larger initial ensembles improve the approximation of the prior uncertainty, but they do not alter the structure of the decision problem. Even with a perfectly specified geological prior, open-loop optimization produces a fixed plan that cannot condition future decisions on observations revealed during execution. In irreversible and long-horizon problems such as mine production scheduling, robustness therefore requires policies that adapt to posterior belief updates, not merely plans that hedge across a static ensemble.

Allowing decisions to depend explicitly on observations fundamentally changes the mathematical nature of the problem. Once learning is acknowledged, the planning task shifts from computing a single optimal schedule to designing a policy: a rule that specifies what action to take given the current state of knowledge about the system. This requires a representation of geological knowledge as a belief state, an explicit observation model describing how actions reveal information, a mechanism for belief updating, and a planning framework that evaluates actions based on both economic return and their effect on future decision quality. Together, these elements place decision-making problems involving geological uncertainty within the class of Partially Observable Markov Decision Processes (POMDPs) [7].

It is important to distinguish the proposed POMDP formulation from model-free reinforcement learning (RL) approaches that have recently been explored in mining applications. While both frameworks involve sequential decision-making, they differ fundamentally in how uncertainty is represented. Model-free RL learns action-value mappings directly from simulated trajectories without maintaining an explicit belief state; decisions are conditioned on observed features rather than on a probabilistic representation of geology. In contrast, the POMDP framework maintains an explicit belief distribution over geological states, updates this belief as observations are obtained, and evaluates actions based on their expected value under the current belief. This distinction is critical: belief-based planning enables principled integration of geostatistical updating with decision-making.

Recent works have demonstrated the potential of POMDP formulations in mineral systems where uncertainty evolves through interaction. For example, POMDP-based approaches have been successfully applied to downstream mineral processing, optimizing flotation operations under feedstock and process uncertainty by explicitly integrating belief updates with control decisions [8]. At a broader strategic level, POMDPs have also been used to manage geological uncertainty in critical mineral supply chains, enabling adaptive sourcing and development decisions in response to imperfect reserve estimates [9].

Beyond mineral systems, POMDPs have demonstrated value over static optimization in diverse domains. In subsurface resource management, sequential decision frameworks have been applied to petroleum reservoir management [10] and drill planning for groundwater management under hydrogeological uncertainty [11]. Medical treatment planning under diagnostic uncertainty has also benefited from belief-based approaches [12]. Recent advances such as BetaZero [13] have extended belief-state planning to long-horizon problems, demonstrating that adaptive decision-making under partial observability can scale to complex sequential tasks previously considered intractable.

Despite these advances, extending POMDP frameworks to upstream strategic mine production scheduling remains an open challenge. Unlike mineral processing or supply-chain sourcing, mine planning involves discrete, combinatorial, and irreversible decisions over three-dimensional block models. Block extraction decisions are constrained by spatial precedence relationships and permanently alter the feasible decision space, leading to a curse of dimensionality that renders direct application of standard POMDP solvers computationally infeasible. Tree-based methods rely on deep rollouts that simulate hypothetical observation sequences far into the future; in the context of mine planning, such rollouts may introduce significant bias and variance in value estimates and quickly become intractable.

As a result, while POMDP formulations have proven effective in mineral systems, mine planning research has largely remained within static or weakly adaptive frameworks focusing on reactive robustness to uncertainty (static hedging) rather than the sequential logic of closed-loop learning. Addressing this gap requires not only adopting a POMDP perspective, but also developing a decision architecture that respects the unique spatial, temporal, and irreversible nature of life-of-mine sequencing.

This paper addresses a different question from existing mine planning literature: "How should block sequencing and cut-off grade decisions be made when mining actions themselves reveal information that alters the optimal future strategy?"

This work does not merely propose a new algorithm for mine planning; it introduces a shift in how the relationship between geological uncertainty, extraction sequencing, and material routing is modeled. Specifically, we make four principal contributions:

i. Formalization of mine production scheduling as a POMDP under irreversibility: We formalize large-scale mine production scheduling as a Partially Observable Markov Decision Process (POMDP), explicitly identifying how geological uncertainty evolves through mining actions and why standard planning formulations cannot condition decisions on this evolution.
ii. A Tractable hybrid simulated annealing (SA)-POMDP architecture via fixed-belief tail optimization: We introduce a hybrid SA-POMDP decision architecture that preserves the sequential feedback structure of a POMDP while remaining computationally tractable.
iii. Decision-centric belief rejuvenation: We adapt existing geostatistical updating techniques- Ensemble smoother-with multiple data assimilation (ES-MDA)- to support belief evolution within a sequential decision framework.
iv. Quantification of the "information axis" in economic valuation: We propose an evaluation framework that distinguishes scenario-dependent outcomes from policy quality.

## 2. Methodology

### 2.1. Mine scheduling as a sequential decision problem

Mine production scheduling is formulated as a sequential decision-making problem under partial observability. At each decision epoch, the planner must select which block to extract and determine its processing destination based on incomplete knowledge of the orebody. Decisions are irreversible and have long-term consequences.

Before introducing the formal framework, it is useful to outline the conceptual structure of this problem in terms familiar to mine planning practitioners. In conventional optimization, the planner commits to a complete extraction sequence before mining begins. This sequence is designed to perform well on average across a set of geological scenarios, but it cannot adapt to what is actually encountered underground. The key insight motivating this work is that mining is not merely an extraction process - it is also a sensing process. Each block extracted reveals grade measurements that carry information about the surrounding geology. A rational planning framework should therefore treat each extraction decision as an opportunity both to generate economic value and to acquire information that improves subsequent decisions.

This dual role of mining actions - producing value while reducing uncertainty - places mine scheduling within the class of sequential decision problems under partial observability. The planner does not know the true geology but maintains a probabilistic representation of what the geology might be (a belief state), updates this representation as observations arrive, and selects actions that are optimal given the current state of knowledge. This is precisely the structure captured by the Partially Observable Markov Decision Process (POMDP) framework.

A POMDP provides a principled mathematical language for problems in which a decision-maker must act sequentially under uncertainty, receiving partial observations that progressively inform its understanding of the system. Importantly, a POMDP separates the formulation of a decision problem from its solution: the formulation defines what must be decided, what is uncertain, and how information is revealed, while solution methods address how to compute approximately optimal decisions under computational constraints. The remainder of this section formalizes mine scheduling within this framework; subsequent sections (2.2-2.3) present a tractable solution method.

#### 2.1.1. Formal POMDP formulation

The mine scheduling problem is defined by the POMDP tuple $< S, A, O, T, R, Z, \gamma >$, where each component is interpreted in the context of open-pit mining as follows:

i. State space $S$: A state $s \in S$ represents the true (but unknown) geological realization combined with the current operational state. The geological component specifies the complete spatial grade distribution across all blocks in the deposit. The operational component records which blocks have been mined, remaining production capacities for each processing stream, and the current time period. Together, these define the full system state at any point during the mine life.

ii. Action space $A$: An action $a \in A$ consists of the joint selection of a feasible block for extraction and its assignment to a processing destination (e.g., sulfide mill, heap leach, or waste dump). Feasibility is governed by spatial precedence constraints - overlying blocks must be removed before deeper blocks become accessible - and by processing capacity limits.

iii. Observation space $O$: Upon extracting a block, the planner observes its true grade values (e.g., copper and gold concentrations). These measurements constitute the observation $o \in O$. Due to spatial correlation in the orebody, a single observation carries information not only about the extracted block but also about unextracted neighboring blocks, enabling the planner to refine expectations about future extraction opportunities.

iv. Transition model $T(s'|s, a)$: The transition function describes how the system state evolves when action $a$ is taken in state $s$. In mine scheduling, transitions are deterministic given the true geological state: extracting a block removes it permanently from the feasible set, adjusts remaining processing capacities for the current period, and advances the operational clock. The geological component of the state does not change - the orebody is fixed; only the planner's knowledge of it evolves.

v. Reward function $R$: The reward captures the immediate economic value generated by extracting block $a$ in state $s$. In this work, the reward corresponds to the discounted net present value (NPV) contribution of the extracted block, which depends on the block's grade, its assigned processing destination, applicable recovery rates, metal prices, and processing costs. Blocks routed to waste destinations generate negative reward (cost of handling) with no metal revenue.

vi. Observation model $Z(o|s', a)$: The observation model specifies how grade measurements are generated after executing action $a$ and transitioning to state $s'$. In this formulation, the extracted block's true grades are observed directly (i.e., the observation is a deterministic function of the true state). The uncertainty arises not from measurement noise but from the fact that the true geological state $s$ is unknown prior to extraction.

vii. Discount factor $\gamma$: In standard POMDP formulations, $\gamma \in (0,1)$ controls the relative weighting of immediate versus future rewards in the Bellman recursion. In the proposed framework, however, decisions are evaluated using a fixed-first-action approximation rather than recursive value iteration (Section 2.2). Temporal preferences are therefore captured directly through the financial discounting embedded in the NPV objective. Accordingly, no separate POMDP discount factor is required, as long-term effects are already reflected in the discounted economic evaluation of continuation plans.

Because the true geological state is not directly observable, the planner cannot condition decisions on state $s$ itself. Instead, decisions are made based on a belief state $b_t$, which represents a probability distribution over geological realizations conditional on all observations obtained up to time $t$. In the ensemble-based representation adopted here, the belief state corresponds to a weighted collection of geological realizations, where the weights reflect each realization's consistency with observed data.

The belief state is the information state of the system: it is a sufficient statistic for decision-making, meaning that all past observations influence future decisions only through the current belief. This property is fundamental - it means that a planner who knows the current belief $b_t$ has exactly as much decision-relevant information as a planner who remembers every individual observation made since mining began.

#### 2.1.2. Value function and belief-based decision-making

With the POMDP components defined, we now describe how the framework evaluates and compares alternative actions. The planner's objective is to select, at each decision epoch, the action that maximizes expected cumulative economic value over the remaining mine life. This objective is expressed through the value function, which assigns to each belief state the expected NPV achievable by following the best available policy from that point onward:

$$V^*(b_t) = max_{a \in A}\, Q^*(b_t, a)$$

where $Q^*(b_t, a)$ is the action-value function, representing the expected long-term economic value of selecting action $a$ when the planner's current state of knowledge is $b_t$. Intuitively, $Q^*(b_t, a)$ answers the question: "If I extract this particular block and send it to this destination, and then act optimally for the rest of the mine life, what is the total expected NPV?"

In a full POMDP formulation, the action-value function accounts not only for the immediate reward but also for how the chosen action will generate an observation, update the belief, and affect all subsequent decisions:

$$Q^*(b_t, a) = E_{s \sim b_t}[R(s, a)] + V^*(b_t')$$

where:

i. $E_{s \sim b_t}[R(s, a)]$ is the expected immediate reward - the belief-weighted NPV contribution of extracting block $a$ across all geological realizations considered plausible under $b_t$.

ii. $V^*(b_t')$ is the optimal value achievable from the updated belief state $b_t'$, obtained after incorporating observation $o$.

The posterior $b_t'$ is obtained by Bayesian conditioning on a simulated observation. For each candidate action $a$, a realization is sampled from $b_t$, the grade that action $a$ would reveal under that realization is simulated, and the belief is conditioned on this simulated observation. The resulting posterior is:

$$b_t'(s') \propto Z(o|s', a) \sum_{s} T(s'|s, a)\, b_t(s)$$

In the present formulation, $T$ acts as the identity on the geological component of the state, so this update reduces to the measurement form:

$$b_t'(s) \propto Z(o|s, a)\, b_t(s)$$

This expression formalizes how the planner revises its probabilistic assessment of the geology: realizations that are consistent with the observed grades receive increased weight, while those that conflict with observations are down-weighted. The hypothetical posterior $b_t'$ is the belief under which the continuation value $V^*(b_t')$ is evaluated, allowing the action-value function to account for the informational effect of $a$ before $a$ is committed. In the ensemble-based implementation adopted in this work, this conditioning is performed by ES-MDA (Section 2.5), which modifies the realizations themselves rather than adjusting explicit probability weights.

#### 2.1.3. From formulation to algorithmic solution

The POMDP formulation presented above defines the structure of the mine scheduling decision problem - what the planner is uncertain about, what actions are available, how observations arise, and what objective is being optimized. It does not prescribe a particular algorithm or solution method.

In principle, solving the POMDP exactly would require computing $V^*(b)$ over the entire continuous belief space, which is intractable for problems of the scale encountered in mine planning. The challenge is therefore to find computationally feasible approximations that preserve the essential structure of belief-based, sequential decision-making while remaining practical for real mining applications.

The following sections address this challenge. Section 2.2 introduces the fixed-first-action approximation that makes belief-based evaluation tractable. Section 2.3 describes the hybrid SA-POMDP decision architecture that combines this approximation with large-scale stochastic optimization. Together, these components provide a practical instantiation of the POMDP framework for strategic mine planning.

### 2.2. Tractability and fixed-first-action approximation

The proposed framework approximates the POMDP solution through a one-step lookahead in belief space. At each decision epoch, the planner evaluates a set of candidate actions as follows. For a given candidate action $a$, a geological realization is sampled from the current belief $b_t$ to generate a simulated grade observation for the considered block. This simulated observation is used to update the belief, producing a posterior belief $b_t'$ that reflects what the planner would know after extracting that block. The remaining mine life is then optimized under this updated belief $b_t'$, yielding a continuation value. The total estimated value of action $a$ is the sum of its immediate reward under $b_t$ and the discounted continuation value under $b_t'$. This procedure is repeated for each candidate action, and the action with the highest estimated value is selected for execution.

Upon execution, the true grade of the selected block is observed- drawn from the hidden true geology rather than from a sampled realization- and the belief is updated accordingly. The planner then proceeds to the next decision epoch with a revised belief that reflects the actual mining observation.

This evaluation structure captures the essential logic of POMDP planning: actions are assessed not only for their immediate economic return but also for how the information they reveal reshapes the geological belief and consequently the quality of all downstream decisions. By updating the belief within the lookahead- before optimizing the continuation- the framework anticipates the informational effect of each action, a mechanism that is absent in receding-horizon approaches where re-optimization operates under a static or passively revised scenario set.

In principle, the accuracy of the estimated action value improves with the number of rollouts: sampling multiple realizations to generate different simulated observations and averaging the resulting continuation values would reduce variance in the Q-value estimates. Strategies such as upper confidence bound (UCB) selection could further balance exploration of uncertain actions against exploitation of promising ones. In the present implementation, a single rollout per candidate action is used at each decision epoch due to the computational cost of repeated life-of-mine optimization. This introduces variance into individual action-value estimates, but the sequential structure of the framework provides a corrective mechanism: because the belief is updated with true observations after each committed action, estimation errors at one epoch do not propagate indefinitely- subsequent re-evaluations under improved beliefs progressively compensate for earlier approximation noise.

### 2.3. Hybrid SA-POMDP decision architecture

The proposed framework combines belief-based decision-making with large-scale stochastic optimization through a hybrid SA-POMDP architecture. Unlike conventional approaches that derive actions implicitly from optimized plans, the proposed framework evaluates candidate actions explicitly through a value function approximation. Each action is assessed by solving a constrained continuation optimization problem, yielding an estimate of its expected long-term impact under the current belief. This defines a mapping from belief states to actions, thereby introducing a policy-based decision structure rather than a plan-driven one.

At each decision epoch:

i. Given $b_t$, a subset of feasible candidate actions is generated.

ii. For each candidate action $a$, a realization is sampled from $b_t$ to generate a simulated observation, and the belief is updated to produce $b_t'$.

iii. The remaining life-of-mine schedule is optimized as a stochastic mine planning problem under the updated belief $b_t'$. This optimization produces a single continuation policy $\pi_a$ that is shared across all geological realizations.

iv. The estimated value of action $a$ is: $Q(b_t, a)$ (detailed calculation steps are provided in supplementary material section 1.)

v. The action $a_t^*$ with the highest estimated value is executed.

vi. The true grade is observed and the belief is updated ($b_{t+1}$).

This architecture replaces explicit belief-space value recursion with repeated stochastic optimization under evolving beliefs. It preserves robustness through shared decisions across scenarios while introducing adaptivity through sequential re-optimization. Fig.1 illustrates the SA-POMDP decision procedure. At each decision epoch, candidate actions are generated (Step 1) and evaluated using the SA-based Q-value oracle (Steps 2a-2d). The action with the highest belief-weighted expected value is selected and executed (Steps 3-4), and the belief state is updated via ES-MDA as new observations are assimilated (Step 5). The following sections detail each component: candidate action selection (Section 2.4), belief updating (Section 2.5), and stochastic optimization via SA (Section 2.6).

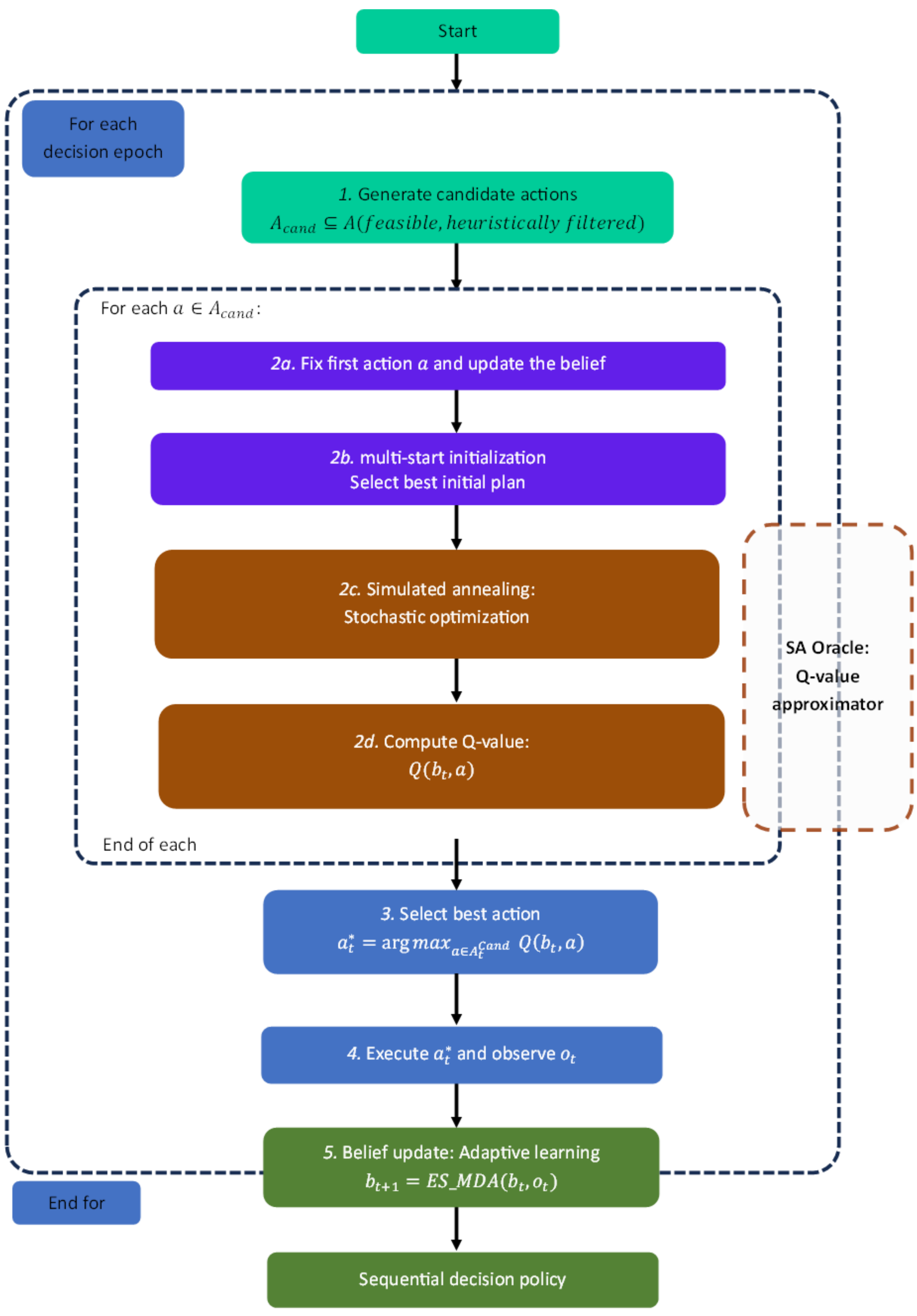

**Fig. 1.** SA-POMDP decision framework for adaptive mine planning under geological uncertainty

### 2.4. Candidate action selection

The feasible action space may contain thousands of blocks at each decision epoch. To ensure tractability, a multi-stage filtering strategy is employed to select a limited set of candidate actions for evaluation.

Candidate selection proceeds as follows:

i. Feasibility filtering: blocks must satisfy precedence constraints.

ii. Heuristic screening: feasible blocks are scored using fast indicators reflecting economic potential, accessibility, and informational relevance.

iii. Diverse sampling to balance exploitation and exploration: a subset of actions is selected to balance economic attractiveness, uncertainty exposure, spatial diversity, and exploration.

Importantly, these heuristics only influence which actions are evaluated, not how actions are ranked. Final selection is based exclusively on belief-weighted long-horizon value estimates. While the framework does not compute the value of information as a separate objective, the one-step lookahead mechanism captures informational effects implicitly: because each candidate action is evaluated under an updated belief that incorporates its simulated observation, actions that reveal more informative grades produce better-calibrated continuation plans and consequently receive higher Q-value estimates. In contrast, one-shot optimization constructs the full extraction sequence ex ante, preventing the assessment of informational effects at the action level.

### 2.5. Belief representation and updating

Geological uncertainty is represented by an ensemble of realizations generated via conditional simulation. Each realization specifies a complete spatial grade distribution and is treated as a plausible state of the world.

Belief updating is performed using the Ensemble Smoother with Multiple Data Assimilation (ES-MDA) [14]. ES-MDA is derived under Gaussian assumptions but has been successfully applied to non-Gaussian settings in reservoir engineering and geosciences, where it serves as a computationally efficient approximation to fully Bayesian updating. As mining observations become available, realizations are updated iteratively using ensemble-based covariance information, ensuring consistency with observed data while preserving spatial structure and ensemble diversity.

ES-MDA avoids particle degeneracy and ensemble collapse, yielding smooth belief evolution suitable for long-horizon sequential decision-making.

(Details of ES-MDA implementation are provided in Supplementary material Section 1).

### 2.6. Action evaluation via simulated annealing

One-shot optimization of the remaining mine life for each candidate action is performed using Simulated Annealing (SA). SA was selected because it has been extensively validated for large scale mine scheduling problems [2,15,16].

For each candidate action, SA solves a stochastic optimization problem in which a single extraction sequence and routing policy are evaluated jointly across all scenarios using belief-weighted aggregation. This ensures that continuation plans are consistent across realizations rather than optimized independently per scenario.

The SA implementation adopts a simplified variant of the algorithm presented in [2], with modifications appropriate for its role as a repeated-call Q-value oracle rather than a standalone global optimizer. Key adaptations include: (i) multi-start initialization from diverse greedy solutions in place of iterative diversification cycles, (ii) reduced iteration counts to ensure computational tractability under repeated evaluation, and (iii) single-temperature geometric cooling rather than adaptive neighborhood-specific temperature schedules.

These simplifications are justified by the oracle's functional requirements: Q-value estimates need only provide accurate relative rankings among candidate actions, not globally optimal solutions. The POMDP framework provides additional robustness through sequential belief updates, which correct for approximation errors as mining observations accumulate.

SA functions as a value oracle within the POMDP framework rather than as a standalone planner.

(Algorithmic details, parameter specifications, and a detailed comparison with [2] are provided in supplementary material Section 2).

## 3. Case study

### 3.1. Mining complex overview

The case study considers a copper-gold open-pit mining complex composed of a single mine supplying material to multiple processing and waste disposal destinations (Fig. 2). The case study is based on geological and operational data provided by a consulting company.

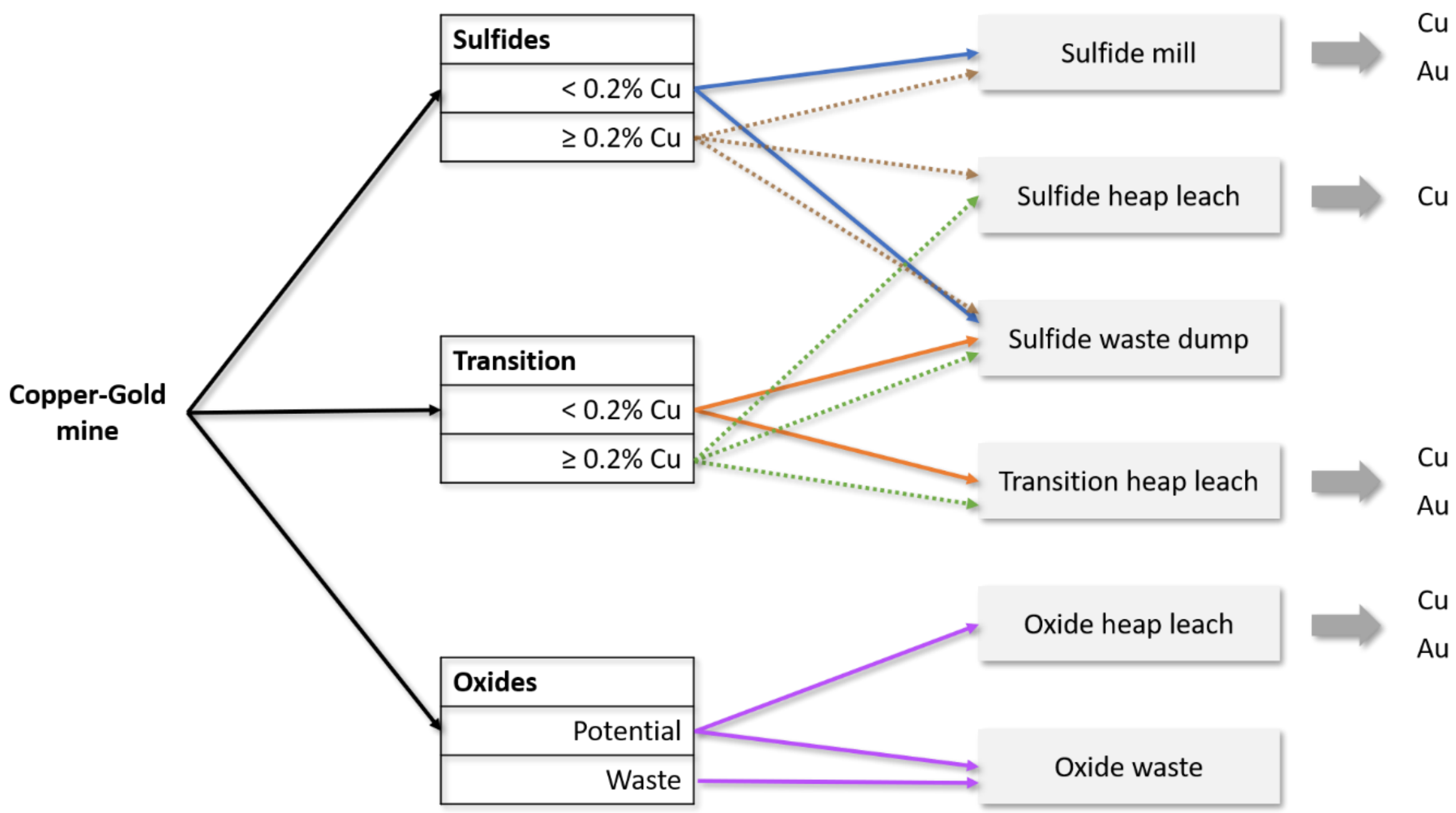

**Fig. 2.** A schematic representation of the mining complex and its associated processing options

The mine contains three primary material groups: Sulfides, transition material, and oxides.

Due to metallurgical and operational constraints, the sulfide and transition materials are further subdivided according to copper grade. Sulfide and transition material with copper grades above or below 0.2 % Cu are treated as distinct material types, reflecting differences in processing suitability. Oxide material is classified as either potentially processable or waste, based on combined copper and gold grade characteristics.

After extraction, material may be routed to one of several destinations, depending on its material type and grade characteristics. The mining complex includes the following processing and disposal options:

i. Sulfide mill, which processes high-grade sulfide material and produces both copper and gold
ii. Sulfide heap leach, which processes sulfide material with lower copper grades
iii. Transition heap leach, which processes transition material above the copper cutoff

iv. Oxide heap leach, which processes oxide material classified as ore

v. Sulfide waste dump, which receives uneconomic sulfide and transition material

vi. Oxide waste dump, which receives uneconomic oxide material

The sulfide mill and sulfide heap leach operate under explicit annual capacity limits, whereas the remaining destinations are assumed to have effectively unlimited capacity (Supplementary material Section 2). Waste destinations do not involve any processing and generate no metal output.

### 3.2. Experimental design and baseline

To isolate the economic value of sequential learning, we compare the proposed SA-POMDP framework against a one-shot stochastic optimization baseline representing conventional mine planning practice. One-shot stochastic optimization is chosen as the baseline because it is the formulation most extensively validated in both academic literature and industrial practice, and therefore provides the most meaningful reference point for practitioners. Rolling-horizon re-optimization with sequential geostatistical updating is a legitimate intermediate baseline that would incorporate belief updating without belief-aware action evaluation. However, such approaches are not yet well established in mine planning practice, and their configuration (re-optimization frequency, updating scheme, horizon truncation) would introduce design choices that confound the comparison. Isolating the marginal value of belief-aware action evaluation from belief updating alone therefore requires a dedicated study, which we defer to future work. Both frameworks are evaluated under identical geological uncertainty, economic parameters, and operational constraints. The experimental design consists of two analyses: (i) a base case under a statistically consistent prior, and (ii) a structured prior misspecification analysis.

#### 3.2.1. Geological ensemble and true geology

An ensemble of 100 conditional geological realizations is generated via sequential Gaussian simulation (SGS) using drillhole data and a common variogram model (Fig. 3). One realization is randomly designated as the hidden true geology for simulation purposes. The remaining 99 realizations form the initial ensemble used for optimization.

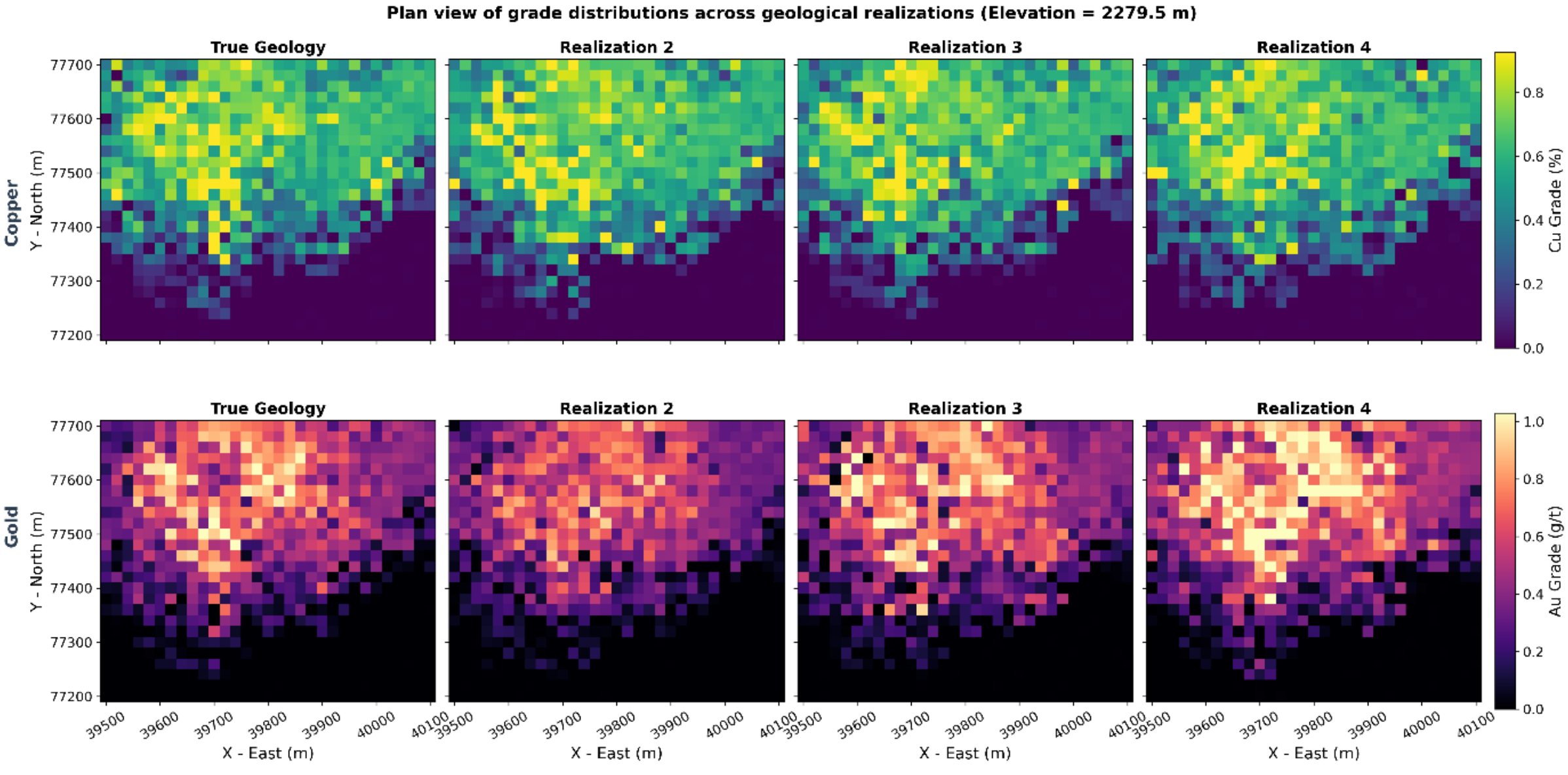

**Fig. 3.** Geological realizations illustrating spatial uncertainty in grade distribution

This design ensures that both frameworks begin with identical information in the base case: neither method has direct knowledge of which realization is true.

#### 3.2.2. Base Case Evaluation (Statistically Consistent Prior)

In the base case ($\alpha = 1.00$), the true realization is drawn from the same statistical model as the ensemble and is excluded from the optimization set.

i. One-shot stochastic optimization receives the 99 realizations and computes a single life-of-mine schedule maximizing expected NPV across scenarios. This schedule is then executed under the hidden true geology.
ii. SA-POMDP with ES-MDA uses the same 99 realizations as its initial belief. At each decision epoch, actions are selected using the fixed-first-action approximation and beliefs are updated sequentially as observations are assimilated.

This base configuration evaluates whether sequential adaptation provides value when the prior model is statistically consistent with reality.

#### 3.2.3. Experimental scope

The base case evaluates the structural value of sequential learning under statistically consistent priors. The misspecification experiments evaluate robustness to systematic mean bias, including both overestimation ($\alpha < 1$) and underestimation ($\alpha > 1$) regimes. Together, these analyses assess the sensitivity of static and adaptive planning to uncertainty magnitude and prior accuracy.

## 4. Results and discussion

### 4.1. Base case outcomes under statistically consistent prior

We first evaluate both frameworks under the base case ($\alpha = 1.00$), where the true geology is drawn from the same SGS prior as the optimization ensemble and excluded from it during planning to ensure out-of-sample evaluation. For each approach, we report the expected NPV under the assumed belief and the realized NPV under the ground-truth geology (Table 1).

**Table 1.** Summary of expected and realized NPVs.

| Method | Expected NPV (M$) | Realized NPV (M$) | Relative Gap |
|---|---|---|---|
| One-shot SA | 221.8 (± 30) | 181.3 | +22.3% overestimation |
| SA-POMDP | 198.5 (± 25.2) | 189.7 | +4.6% overestimation |

Although the one-shot approach achieves the highest expected NPV, its realized performance under the hidden true geology is substantially lower, yielding a +22.3% optimism gap. The SA-POMDP reduces this gap to +4.6%, indicating substantially improved alignment between expectation and execution.

The base case reports realized NPVs under a single hidden true realization drawn from the prior ensemble. This single-draw design is sufficient to demonstrate that the expectation-reality gap can differ structurally between frameworks, but it does not characterize the distribution of realized outcomes across alternative true geologies. The magnitude of the gap reduction reported here should therefore be interpreted as indicative rather than as a population estimate. The prior misspecification analysis (Section "Robustness under prior misspecification") provides complementary evidence that the qualitative direction of the effect - adaptive planning narrowing the gap more than static planning - is stable across five independent true geologies corresponding to different bias regimes. A systematic Monte Carlo evaluation across many true realizations per regime is a natural extension and is discussed in Section "Future research directions".

Fig. 4 presents the cumulative discounted cash-flow profiles for both frameworks, and as mining progresses, the divergence becomes pronounced. The one-shot true profile consistently tracks the lower bound of its uncertainty envelope, indicating that early sequencing commitments amplify the deviations. The adaptive framework limits this amplification: the POMDP true profile remains close to the central tendency of the evolving belief distribution, suggesting that sequential re-optimization maintains calibration even as decisions become spatially sensitive.

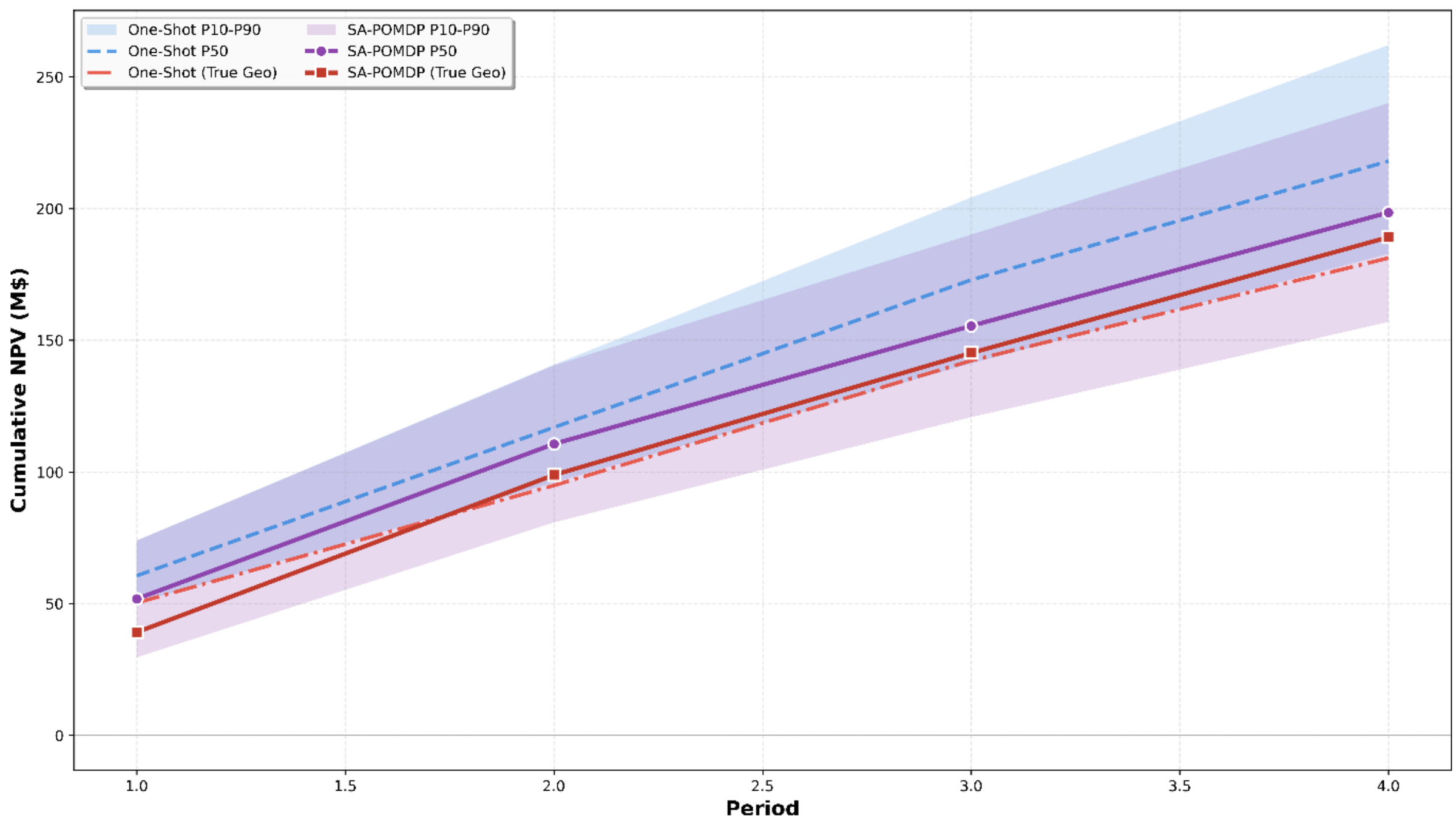

**Fig. 4.** Cumulative discounted cash-flow profiles

### 4.2. Operational mechanisms: material flow and grade dynamics

Fig. 5 compares annual material flows across processing streams under both frameworks. Under static planning, the realized flows indicate limited structural flexibility. Because sequencing decisions are committed ex ante, deviations from expected spatial patterns manifest as local routing adjustments rather than structural corrections. The SA-POMDP framework produces a structurally distinct operating regime: by updating beliefs sequentially and re-evaluating actions under revised uncertainty, the adaptive policy reallocates material flows dynamically.

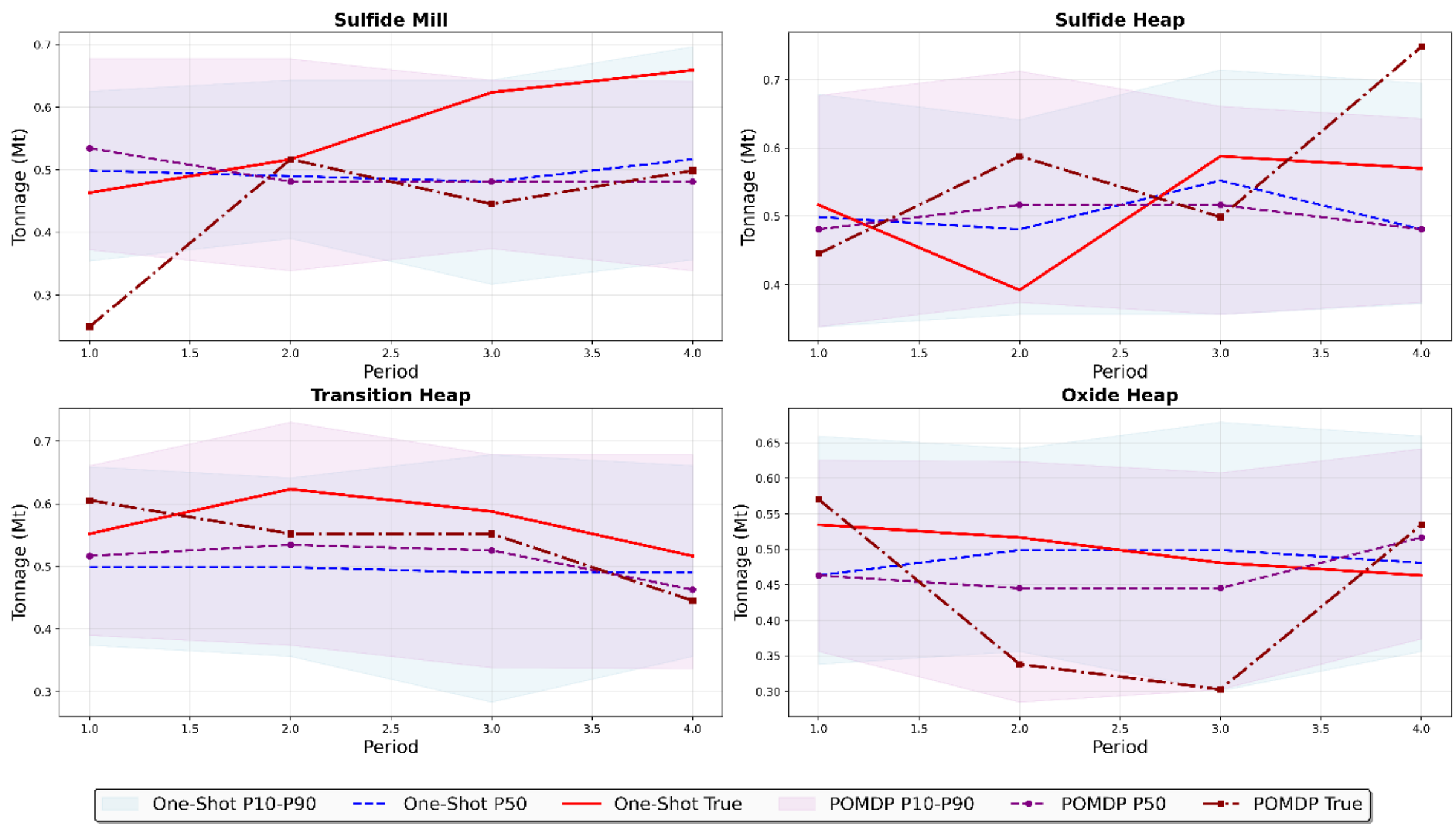

**Fig. 5.** Annual material flow comparison.

Fig. 6 and Fig. 7 present the annual average Cu and Au grades by destination, with uncertainty envelopes and realized trajectories. Several patterns emerge:

i. The one-shot optimizer maintains relatively smooth expected grade profiles, but realized grades under the true geology deviate from expectations in multiple destinations

ii. The SA-POMDP grade trajectories adjust dynamically and tend to remain closer to the P50.

iii. Differences in grade behaviour are not isolated to a single stream but reflect coordinated adjustments across sulfide and oxide destinations.

These differences are also evident in spatial extraction sequences and destination maps as shown in Fig.8 and Fig.9.

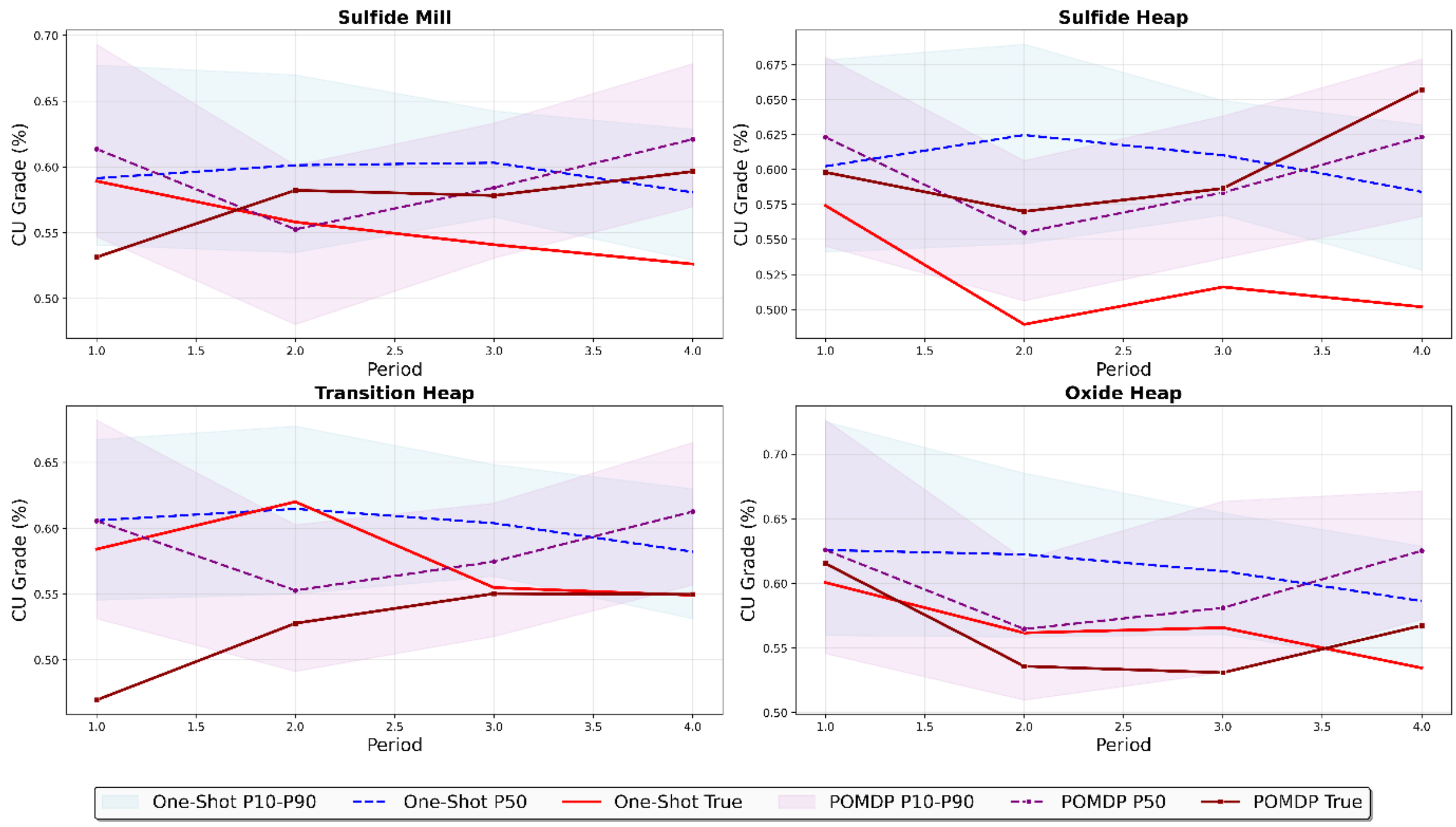

**Fig. 6.** Cu average grade profiles

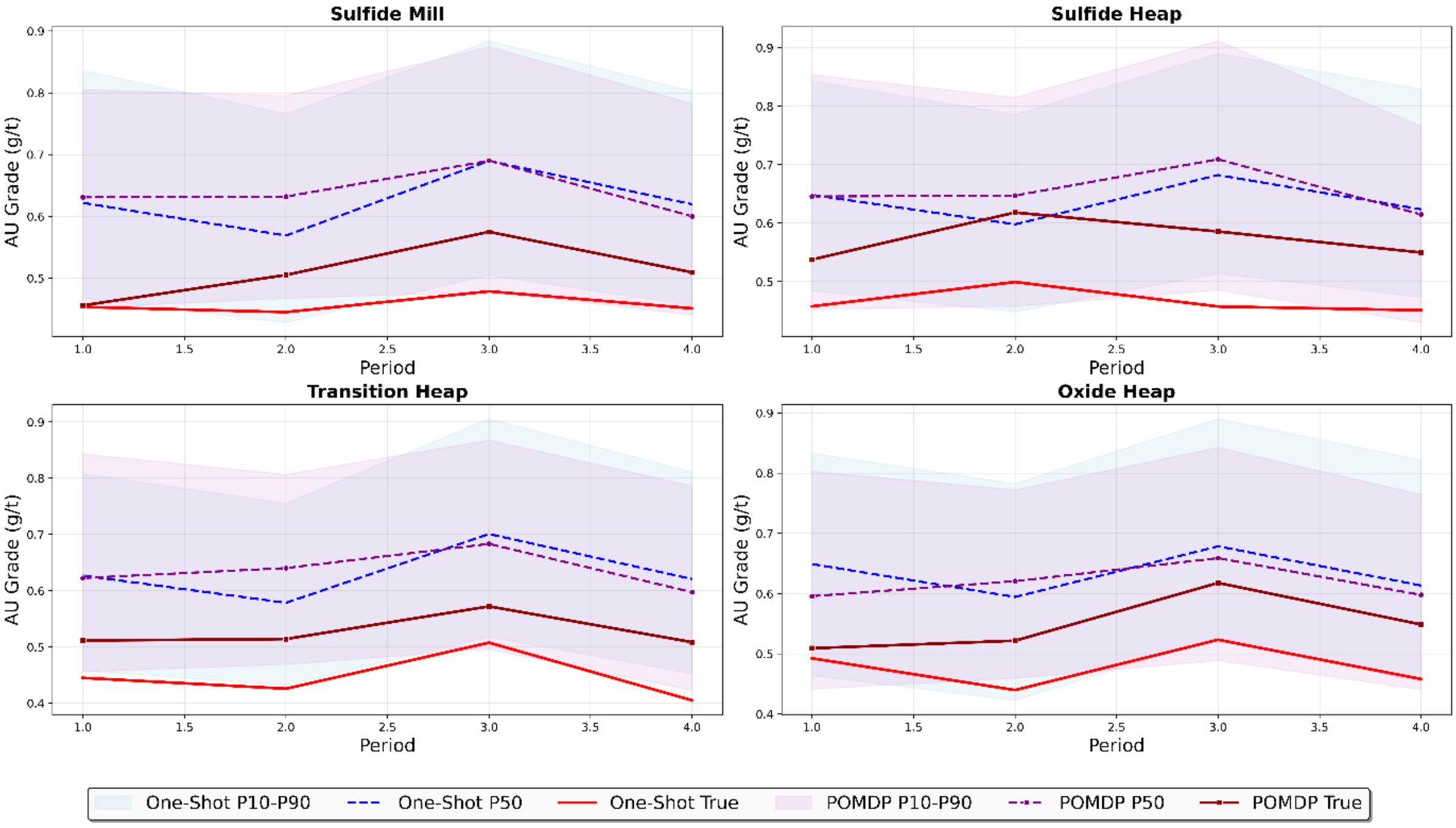

**Fig. 7.** Au average grade profiles

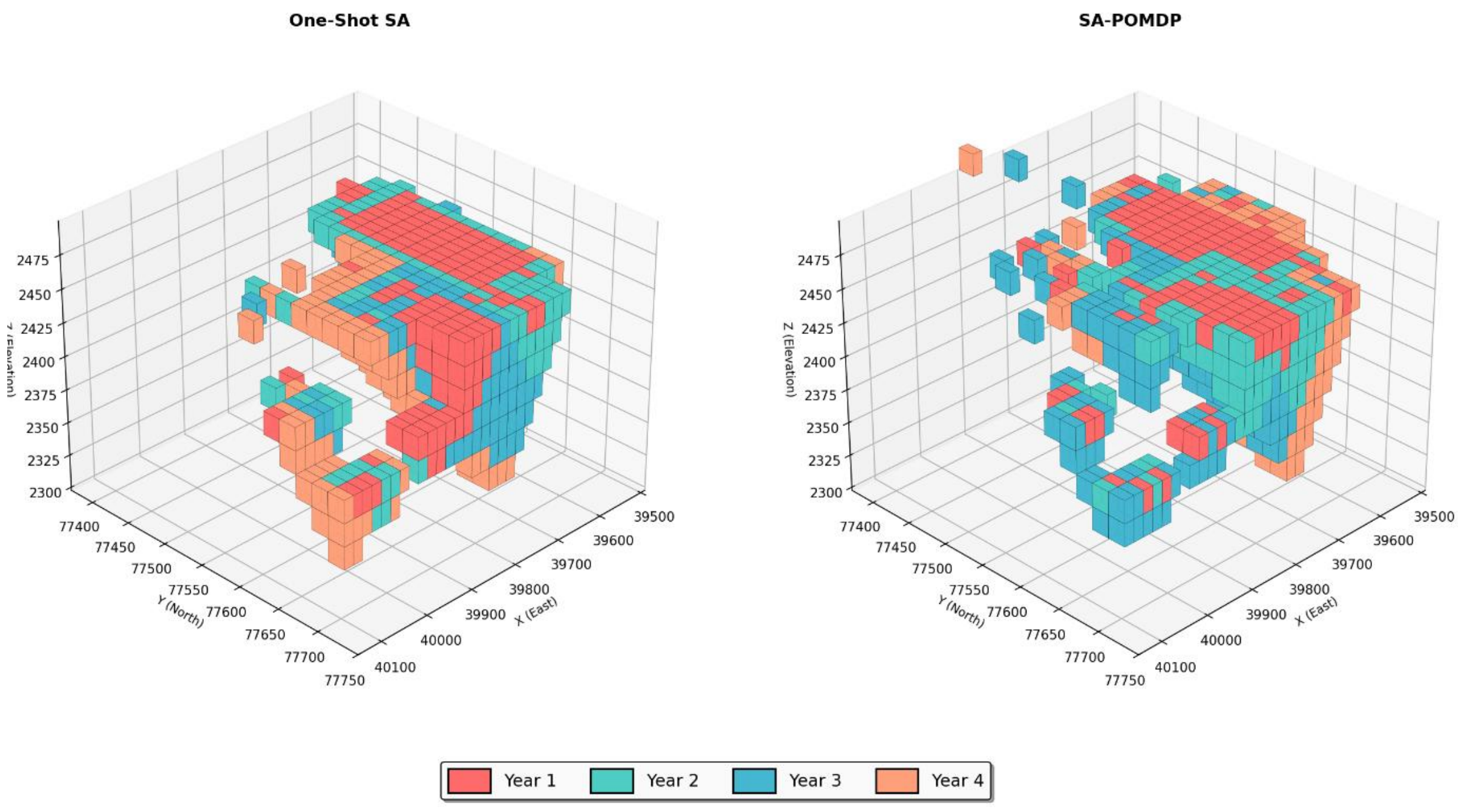

**Fig. 8.** Spatial extraction sequences for the one-shot and SA-POMDP frameworks over years.

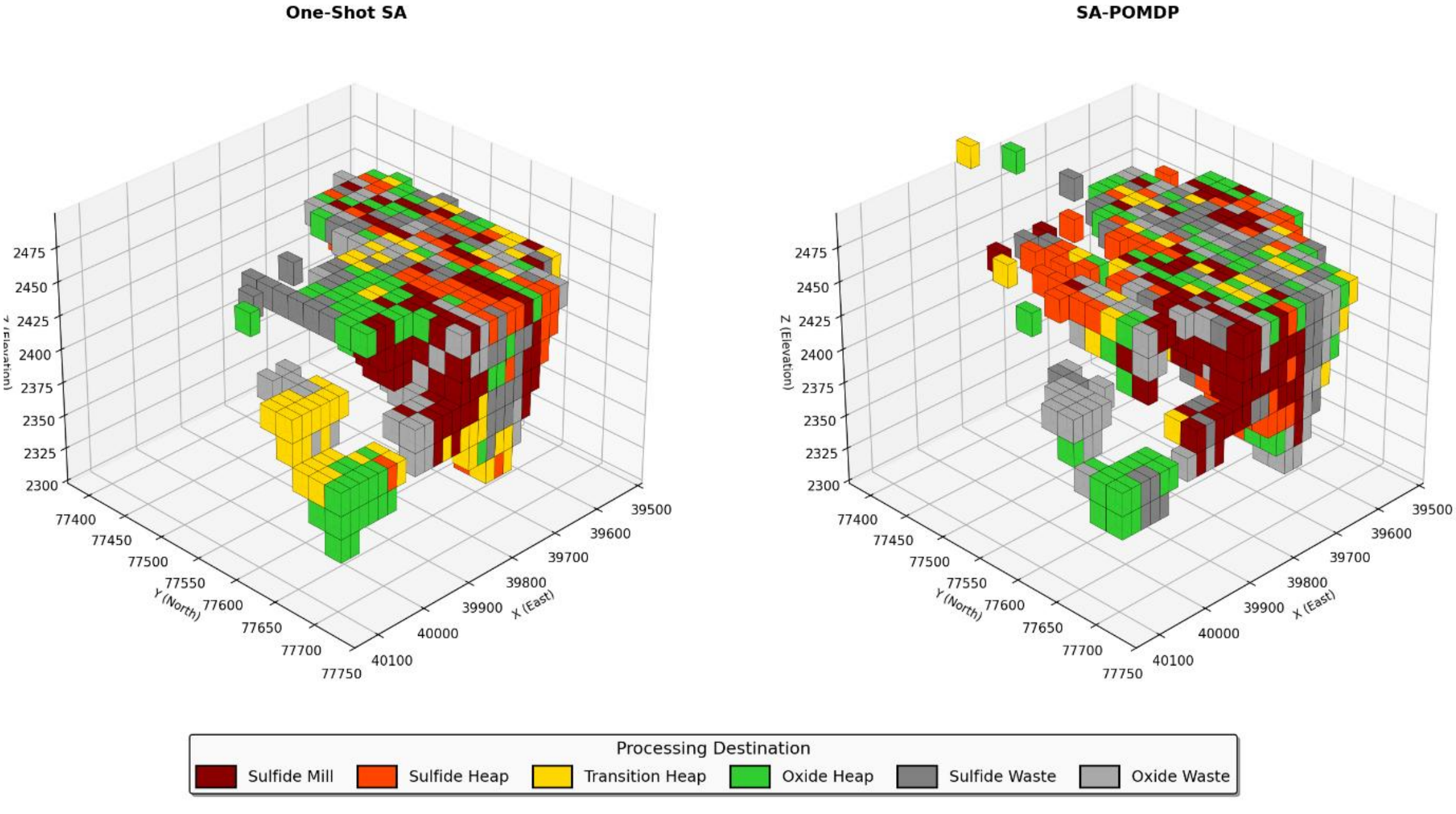

**Fig. 9.** Spatial distribution of processing destinations for extracted material under the one-shot and SA-POMDP frameworks

### 4.3. Belief dynamics

Fig. 10 presents the evolution of ensemble uncertainty under ES-MDA updating. The mean spread represents average block-support standard deviation across realizations, while the shaded band denotes the spatial dispersion of that uncertainty.

During early assimilation, absolute spread increases before stabilizing. This reflects variance restructuring under sequential conditioning. Receiving observations and updating the belief may increase spread as realizations are corrected: those that over-predict observed grades are adjusted downward, while those that under-predict are adjusted upward, causing ensemble disagreement to grow.

After approximately 80-100 mining steps, both mean spread and its spatial dispersion stabilize, indicating convergence to a posterior uncertainty regime under partial observability. Convergence here is distributional rather than pathwise: realizations remain diverse, but their collective variance structure stabilizes.

Fig. 11 shows the coefficient of variation of the spread (CV-spread), defined as the spatial standard deviation of spread divided by its mean. The decline and stabilization of CV-spread suggest increasing structural consistency in the spatial distribution of uncertainty.

The stabilization of CV-spread at approximately 30% indicates that the spatial structure of uncertainty has converged- not that all blocks are equally certain, but that the pattern of uncertainty now respects the deposit's spatial correlation structure and the distribution of observations. This calibrated uncertainty regime ensures that subsequent sequencing and routing decisions are evaluated under a realistic representation of remaining geological risk.

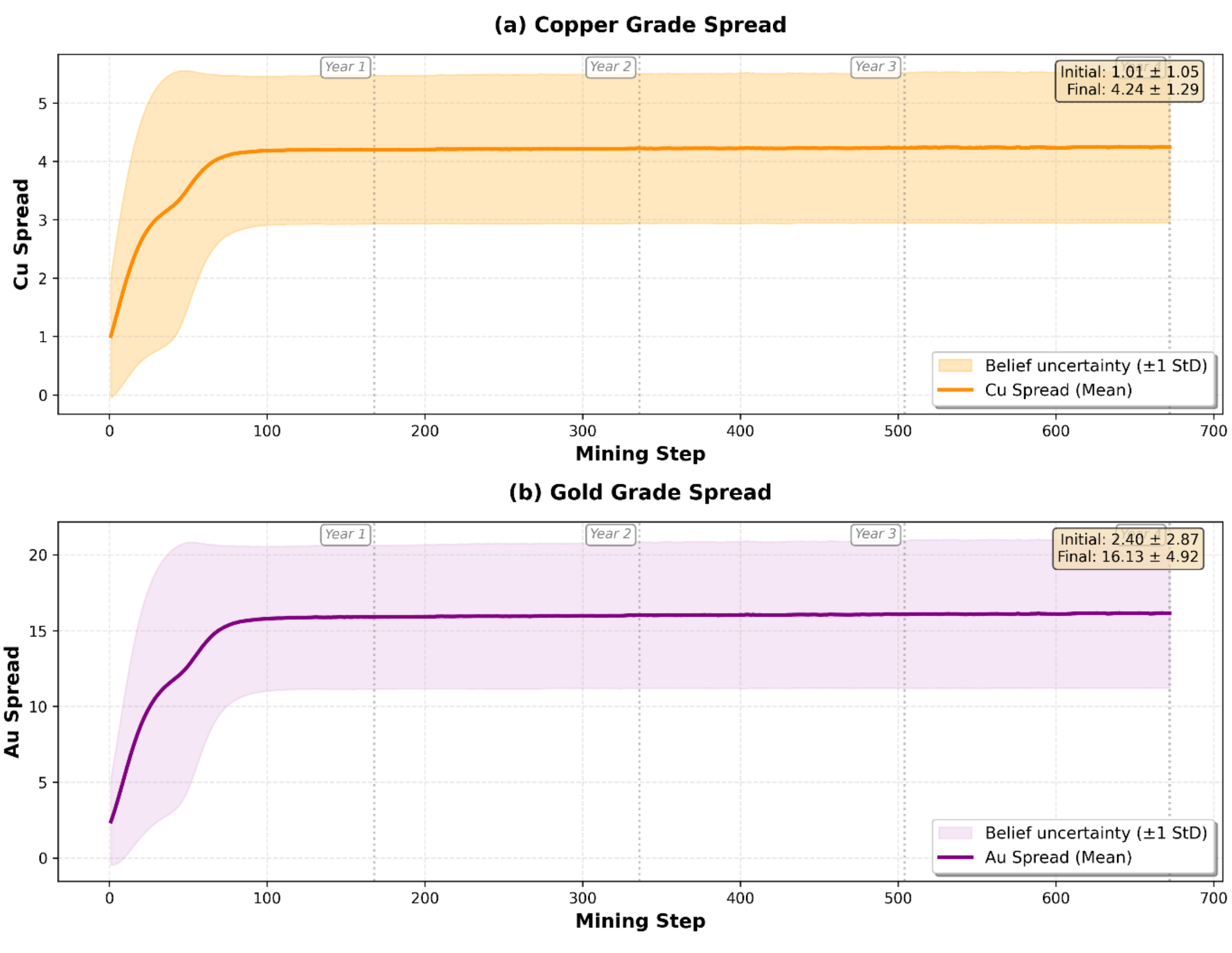

**Fig. 10.** Cu and Au spread evolution

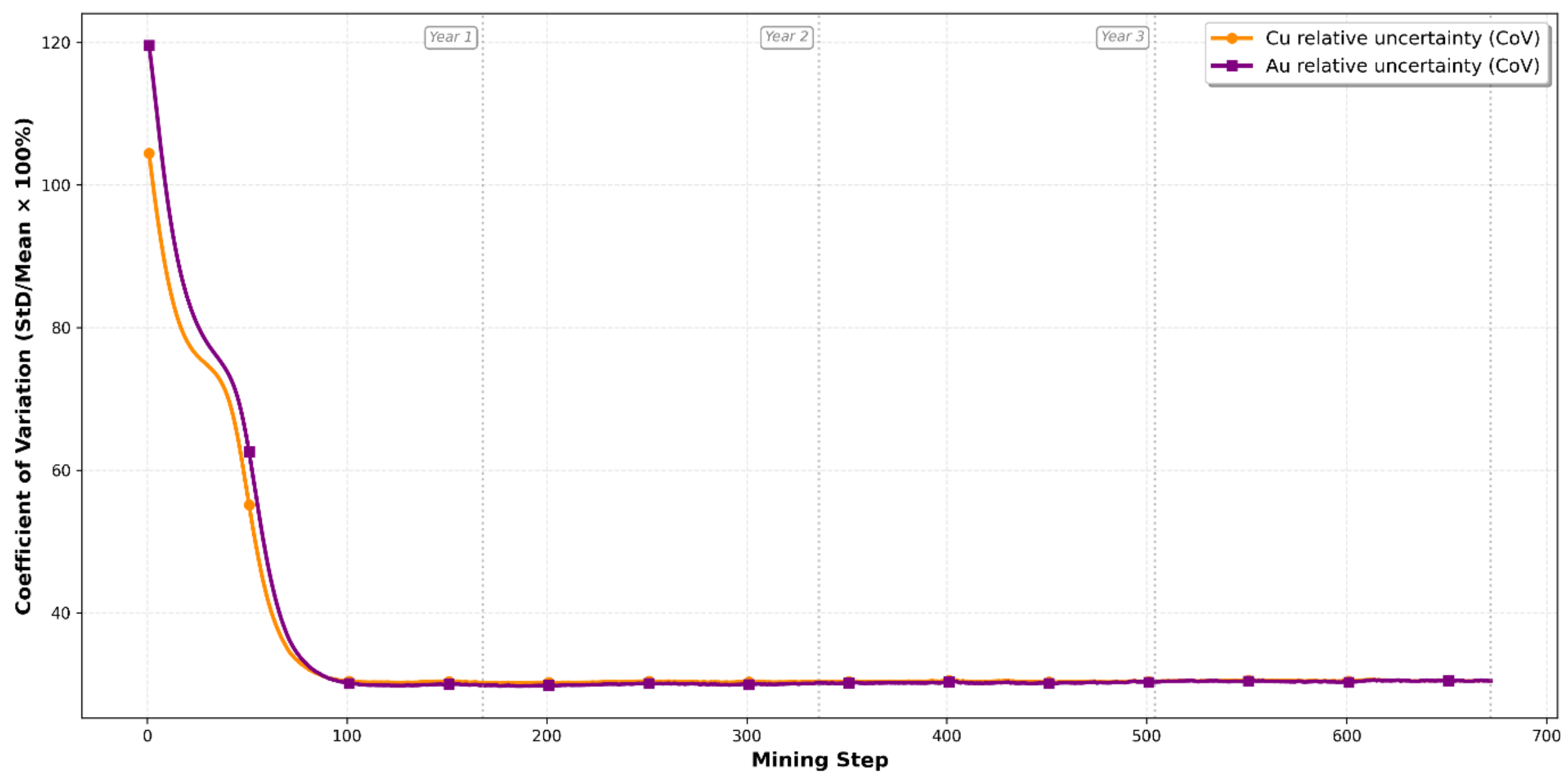

**Fig. 11.** Evolution of Cu and Au coefficients of variation of the spread

### 4.4. Robustness under prior misspecification

To evaluate structural robustness, we conducted a controlled prior misspecification analysis by applying multiplicative scaling factors $\alpha \in \{0.90, 0.95, 1.05, 1.10\}$ (corresponding to -10%, -5%, +5%, +10% bias) to the true geology while holding the initial optimization ensemble fixed.

Fig. 12 summarizes expected and realized NPVs across regimes. Several patterns emerge:

i. Under downward bias ($\alpha < 1$), realized NPV for the one-shot optimizer deteriorates sharply as systematic overestimation increases.
ii. The SA-POMDP degrades more gradually, demonstrating reduced downside amplification.
iii. Under upward bias ($\alpha > 1$), both methods improve, but the adaptive framework captures upside more effectively.

Importantly, in misspecification experiments the scaled true realization is included within the one-shot optimization ensemble. Thus, performance differences reflect structural decision dynamics rather than absence of representational support. Even when the true geology is representable within the scenario set, static open-loop optimization remains sensitive to bias magnitude.

The performance gap widens as bias magnitude increases, demonstrating that adaptive policies scale more gracefully under systematic deviation. For example, at $\alpha$=0.90, the POMDP outperforms one-shot by 44.6 M$ (36.9%); at $\alpha$=1.10, the advantage is 24.1 M$ (11.7%).

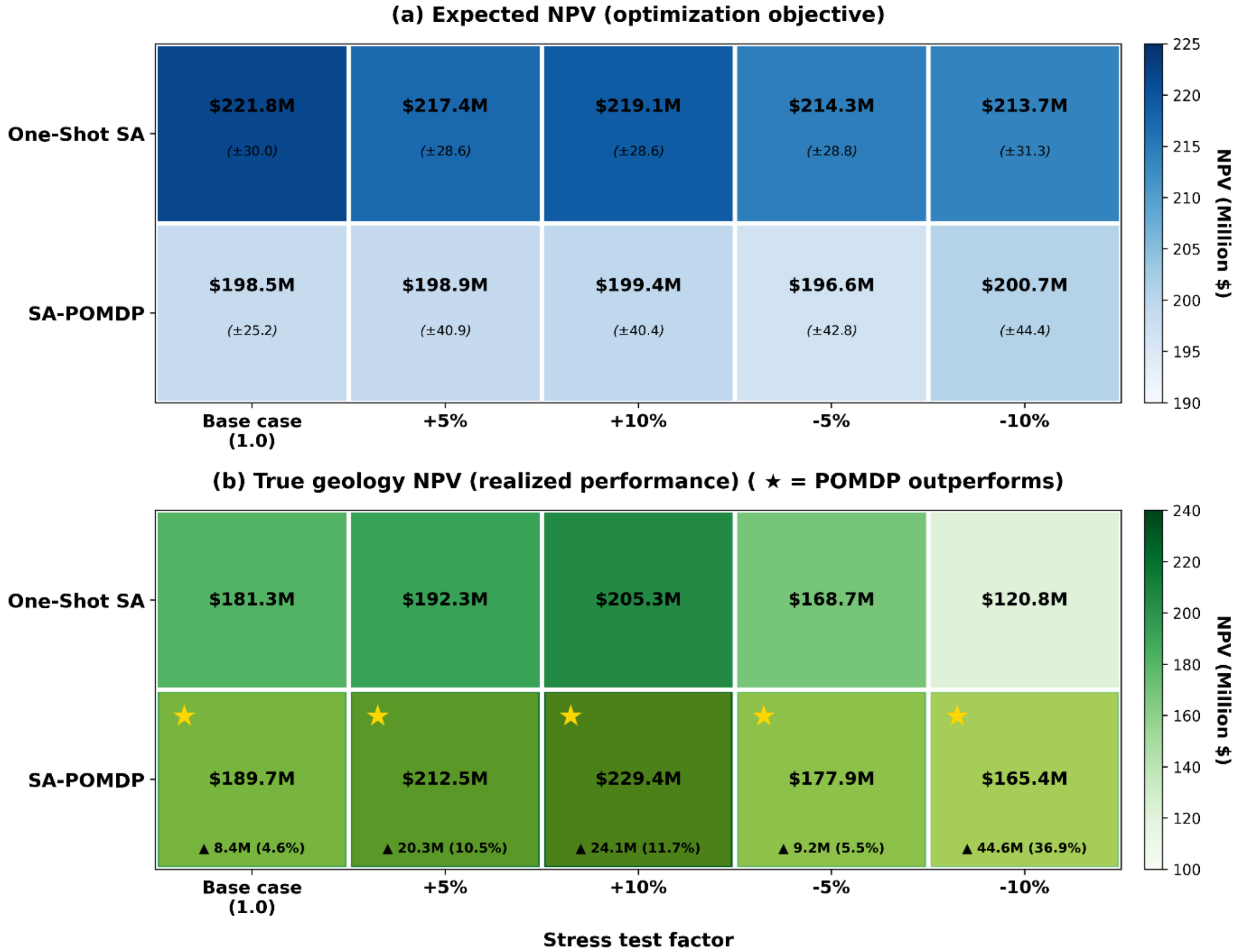

**Fig. 12.** Expected and realized NPVs under different misspecified priors

### 4.5. Risk analysis: From static robustness to dynamic risk management

Conventional stochastic mine planning evaluates risk by re-scoring the optimized plan across the same ensemble of geological realizations used during optimization. This leads to static robustness, which is internally consistent but potentially over-optimistic when the prior is misspecified.

Such an approach is meaningful only if the initial ensemble remains statistically consistent with reality throughout execution. In early project stages, however, ensembles are often conditioned on drillhole data. In addition, even if the true geology lies in the ensemble, the one-shot optimizer has no mechanism to recognize it. Risk management therefore reduces to hedging across scenarios rather than adapting decisions as information is revealed.

The SA-POMDP framework replaces static robustness with dynamic risk management. Risk is no longer evaluated by the width of an ex-ante NPV distribution, but by the stability of realized performance under sequential belief updates. As observations recalibrate the belief state, decisions are re-optimized under an uncertainty structure that increasingly reflects true spatial heterogeneity.

Crucially, this recalibration does not eliminate uncertainty. Instead, it enables the decision-maker to distinguish actions that remain viable across the full range of plausible outcomes from those that are fragile under modest variations. This implicit filtering explains why the SA-POMDP yields lower expected NPV but substantially higher realized NPV.

The value of learning is regime-dependent: as uncertainty magnitude and bias increase, adaptive advantage becomes more pronounced. In low-uncertainty regimes, performance differences narrow, consistent with decision-theoretic expectations.

### 4.6. The information axis and conservative optimality

The previous results support the central thesis of this work: time is an information axis, not merely a discounting index. The one-shot optimizer maximizes expected value under the initial ensemble, but cannot adjust when subsequent observations alter the effective uncertainty structure.

The SA-POMDP framework does not assume access to future information when evaluating actions. Instead, decisions are made based on the current belief state, and future information is incorporated only as it is observed through mining. Learning therefore influences performance sequentially through belief updates. This explains why the POMDP's realized performance remains centered in its uncertainty envelope (Fig. 4)- decisions adapt as information arrives rather than committing prematurely.

### 4.7. Implications for mine planning practice

These findings indicate that one-shot stochastic optimization is not incorrect, but structurally incomplete for problems where actions reveal information and irreversibly constrain future options. Periodic re-optimization remains reactive, as each re-solve is anchored to a static view of the remaining life-of-mine.

The SA-POMDP with ES-MDA represents a proactive learning paradigm. Rather than repeatedly recalculating plans, it updates the underlying belief and decision logic. In doing so, it reduces dependence on scenario luck and increases reliance on policy adaptability, transforming uncertainty from a passive risk into an active driver of robust decision-making.

## 5. Future research directions

While the proposed SA-POMDP framework demonstrates the structural advantages of adaptive mine planning under epistemic uncertainty, several important research directions remain open.

### 5.1. Statistical validation and baseline extension

Two empirical extensions are priorities for follow-on work. First, Monte Carlo evaluation across many independently sampled true geologies per uncertainty regime would replace point estimates of the expectation-reality gap with proper distributions, enabling confidence intervals on the adaptive advantage and sensitivity analysis with respect to the specific realization used. Second, comparison against rolling-horizon re-optimization with sequential geostatistical updating would isolate the marginal value of belief-aware action evaluation from belief updating alone, clarifying which portion of the adaptive advantage reported here derives from the POMDP structure versus from the information assimilation mechanism. Together, these extensions would move the framework from a paradigm demonstration to a quantitatively characterized decision method.

### 5.2. Computational scalability

The adaptive structure of the SA-POMDP framework incurs substantially higher computational cost than one-shot optimization. In the present case study, the SA-POMDP required approximately 60 CPU-hours compared to few minutes for the one-shot baseline. A promising direction includes learning-based value approximators trained on simulated scheduling outcomes. The structure of the proposed framework naturally generates state-action-value samples of the form $(b_t, a, Q(b_t, a))$, providing a foundation for learning-based policy approximation. This opens the possibility of replacing repeated SA-based optimization with surrogate models that approximate long-horizon action values, enabling faster decision-making and potential real-time applications.

### 5.3. When is learning economically valuable?

The results show that adaptive planning reduces the expectation-reality gap under significant geological uncertainty. As uncertainty decreases or prior models become highly accurate, the marginal benefit of sequential adaptation diminishes. Conversely, as epistemic uncertainty and spatial heterogeneity increase, the economic amplification of irreversible sequencing errors becomes more pronounced. A systematic investigation of this regime sensitivity is essential. Future research should quantify the relationship between uncertainty magnitude and the economic value of learning. Such analysis would transform the framework from a case-study demonstration into a predictive tool for determining when adaptive planning is economically justified.

### 5.4. Integration with short-term planning

The present framework operates at the strategic planning level, focusing on long-term sequencing and routing decisions. In practice, however, mine operations function with a hierarchical structure. Short-term operational constraints- equipment availability, blending constraints, stockpile management, and real-time plant performance- are not explicitly modeled in the current formulation.

Future research should explore hierarchical or multi-agent extensions of the SA-POMDP framework in which:

i. A strategic-level adaptive planner communicates uncertainty informed targets,
ii. Tactical and operational agents respond to real-time constraints,
iii. Information flows bidirectionally across planning horizons.

Such multi-layered adaptive architectures would enable real-time adaptive decision-making while preserving long-term strategic coherence. This represents a shift from static hierarchical planning to coordinated adaptive control across temporal scales.

## 6. Conclusion

This work introduces a fundamentally different perspective on strategic mine planning under geological uncertainty by formulating the problem as a sequential decision process rather than a static optimization task. The proposed SA-POMDP framework enables explicit action-level evaluation under evolving beliefs, shifting the focus from generating fixed plans to designing adaptive policies that map belief states to actions.

The empirical results demonstrate that this policy-based approach substantially improves the alignment between expected and realized economic performance. Under a statistically consistent prior, the framework reduces the expectation-reality gap from 22.3% to 4.6%. Under prior misspecification, the adaptive advantage becomes more pronounced: when the true geology deviates by 10% from the assumed prior, the SA-POMDP outperforms static optimization by up to 37% in realized NPV. These findings confirm that sequential belief updating provides structural robustness that static hedging across scenarios cannot achieve.

Importantly, the proposed method does not explicitly optimize the value of information, but captures its effects implicitly through sequential action evaluation and belief updating. This provides a computationally tractable alternative to exact and tree-based POMDP solvers while retaining the key benefits of adaptive decision-making.

Beyond performance improvements, the primary contribution of this work is conceptual: it reframes strategic mine planning as a policy design problem under partial observability. This perspective opens new avenues for integrating learning-based value approximations, quantifying the economic value of information across uncertainty regimes, and extending the framework to hierarchical multi-scale planning architectures. By treating time as an information axis rather than merely a discounting index, the framework transforms geological uncertainty from a passive risk into an active driver of robust decision-making.

**Acknowledgment**

This work was done at Geology and Sustainable Mining Institute (GSMI) under the special agreement OCP-UM6P APRA project.

**Statement and declarations**

The authors declare that they have no known competing financial interests or personal relationships that could have appeared to influence the work reported in this paper.